# Enhanced Genetic Programming Models with Multiple Equations for Accurate Semi-Autogenous Grinding Mill Throughput Prediction

Zahra Ghasemi
*School of Electrical and Mechanical Engineering*
*The University of Adelaide*
Adelaide, Australia
zahra.ghasemi@adelaide.edu.au

Mehdi Neshat
*Centre for Artificial Intelligence Research & Optimisation*
*Torrens University*
Brisbane, Australia
mehdi.neshat@torrens.edu.au

Chris Aldrich
*Western Australian School of Mines*
*Curtin University*
Perth, Australia
chris.aldrich@curtin.edu.au

John Karageorgos
*Manta Controls Pty Ltd*
Adelaide, Australia
john.karageorgos@mantacontrols.com.au

Max Zanin
*School of Chemical Engineering*
*The University of Adelaide*
Adelaide, Australia
max.zanin@mzminerals.com

Frank Neumann
*School of Computer and Mathematical Sciences*
*The University of Adelaide*
Adelaide, Australia
frank.neumann@adelaide.edu.au

Lei Chen
*School of Electrical and Mechanical Engineering*
*The University of Adelaide*
Adelaide, Australia
lei.chen@adelaide.edu.au

*Abstract*—Semi-autogenous grinding (SAG) mills play a pivotal role in the grinding circuit of mineral processing plants. Accurate prediction of SAG mill throughput as a crucial performance metric is of utmost importance. The potential of applying genetic programming (GP) for this purpose has yet to be thoroughly investigated. This study introduces an enhanced GP approach entitled multi-equation GP (MEGP) for more accurate prediction of SAG mill throughput. In the new proposed method multiple equations, each accurately predicting mill throughput for specific clusters of training data are extracted. These equations are then employed to predict mill throughput for test data using various approaches. To assess the effect of distance measures, four different distance measures are employed in MEGP method. Comparative analysis reveals that the best MEGP approach achieves an average improvement of 10.74% in prediction accuracy compared with standard GP. In this approach all extracted equations are utilized and both the number of data points in each data cluster and the distance to clusters are incorporated for calculating the final prediction.
Further investigation of distance measures indicates that among four different metrics employed including Euclidean, Manhattan, Chebyshev, and Cosine distance, the Euclidean distance measure yields the most accurate results for the majority of data splits.

## I. Introduction

In mineral processing plants, the separation of valuable minerals from the waste material involves various steps. Among these steps, grinding is a pivotal step, accounting for approximately 50% of mineral processing costs [1]. Semi-autogenous grinding (SAG) mills are extensively utilised in the grinding circuit of mineral processing plants. These huge pieces of equipment utilise both steel balls and coarse ore for size reduction, distinguishing them from autogenous grinding (AG) mills, which solely rely on the ore for grinding [2].

SAG mill throughput is a crucial performance metric indicating the amount of processed ore per hour [3]. Precise prediction of SAG mill throughput is crucial for informed decision making and process optimisation. These predictive models enable the examination of different set points for controllable parameters to identify the most appropriate scenario. Moreover, reliable throughput prediction models provide insights into the overall impact of input variables on mill throughput. Furthermore, when integrated with optimisation algorithms, these models have the capability to recommend optimal input parameters, ultimately leading to the maximisation of mill throughput.

Extraction of empirical equations for mill throughput is focused in some publications [4, 5]. These models are based on Bond's equation [6] and Morrel models [7-9] for calculating the specific energy of grinding mills. The specific energy is the amount of energy required to reduce the size of ore to a specific particle size. The empirical models for mill throughput prediction exhibit accurate results. However, obtaining these equations requires conducting numerous controlled experiments, which are time-consuming and expensive. Furthermore, their applicability remains restricted to the specific experimental conditions employed. In other words, their findings cannot be generalised to other grinding circuits. Even for the same grinding circuit, the accuracy of the derived model may decline over time due to factors such as equipment wear and changes in ore hardness. Consequently, these empirical models, which necessitate substantial time and resources, have limited long-term validity even within the same mining plant.

This research was supported by the Australian Research Council Integrated Operations for Complex Resources Industrial Transformation Training Centre (project number IC190100017).

Having discussed the challenges of empirical modelling, and taking into account the huge amount of production data stored in the databases of mineral processing plants, machine learning techniques can be applied to utilise this enormous amount of data and provide accurate prediction models [10]. Numerous machine learning models can be utilised for this purpose, such as recurrent neural network [11], genetic programming [12], support vector regression [13], and random forest regression [14], to name a few.

Among different prediction models, genetic programming (GP) is a potent technique that employs genetic algorithms to derive an explicit mathematical relationship between input and output variables. GP is able to discover complex and non-linear relationships within the data. The evolutionary nature of GP enables it to adapt and refine the predictive model over successive generations, continuously optimising its performance. This inherent adaptability is particularly advantageous when dealing with the intricate and dynamic nature of SAG mill processes. Furthermore, GP excels in transparency, providing interpretable equations that offer insights into the underlying relationships within the data. This transparency is a key asset, especially in industries such as mineral processing, where the interpretability of models is often as important as predictive accuracy. Since the development of GP by Koza, this method attracted a lot of interest in modelling numerous real-world problems. For instance, Chen et al. [15] introduced a dynamic heuristic algorithm for container terminal truck dispatching, utilising manual and GP heuristics. Results demonstrate that both heuristics enhance operational efficiency, with GP heuristics exhibiting a substantial 59% improvement over the practical method. In another study human activity classification using CW Doppler radar returns, employing computational intelligence techniques is investigated [16]. Genetic programming models demonstrated high accuracy and explainability, even outperforming established black-box methods, showcasing their effectiveness in producing concise and insightful models for radar-based human activity classification. Moreover, a diverse range of other applications exists, including job shop scheduling [17, 18], traffic modelling [19], and financial forecasting problem [20].

GP has been also applied in processing. McKlay et al. [21] utilized GP for modelling chemical processes using plant data. In another study, GP was used to model plasticising extrusion process [22]. Recurrent neural network (RNN) and GP were employed for this objective, and comparative analyses demonstrated that GP outperformed RNN significantly in the obtained results. In another study GP has been utilised to develop dynamic models for predicting the quality of extruded food products having process inputs [23]. For this case GP outperformed feedforward artificial neural network (ANN). Another application is for leaching experiments including acid pressure leaching of nickeliferous chromite's and leaching of uranium and radium [24]. In this case GP results were either of comparable accuracy or significantly more accurate than regression models. Ross et al. [25] used GP to evolve mineral identifiers for hyperspectral images, enabling the automatic classification of specific minerals based on their spectral signatures. Feng et al. [26] combined GP and an improved particle swarm optimisation algorithm to simultaneously establish the structure and parameters of a visco-elastic rock material model. The obtained results indicates that the proposed hybrid evolutionary algorithm is effective in achieving global optimum recognition for both the nonlinear visco-elastic material model structure and its coefficients. Hoseinian et al. [27] used gene expression programming (GEP) to model the SAG mill power consumption. GEP could outperform non-linear multiple regression (NLMR) method in both prediction error and correlation coefficient. Shirani Faradonbeh et al. [28] employed GP and GEP for predicting the performance of road heading machines and GEP outperformed GP. Qi et al. [29] developed a prediction model for uniaxial compressive strength (UCS) cemented paste backfill using GP. The developed GP model demonstrated superior performance over random forest (RF) and decision trees (DT), while also yielding comparable results to the gradient boosting machine (GBM).

The objective of this research is applying and improving GP for predicting SAG mill throughput. In this study, a new GP method named multi-equation GP (MEGP) is proposed. The underlying principle behind MEGP is to generate multiple equations for mill throughput prediction, extracting subgroups of train data that exhibit a better fit with each extracted equation, and then utilising these equations to calculate predictions for test data using different approaches. MEGP will be implemented using four different distance measures including Euclidean, Manhattan, Chebyshev, and Cosine distance to assess the impact of these measures on model's performance. By considering a variety of distance measures in developing MEGP, experts can adapt the evaluation process to the specific problem domain. The choice of distance measures should be driven by the characteristics of the data being used. The following research questions will be addressed in this study:

RQ1. Which MEGP prediction approach is the most successful for predicting SAG mill throughput?

RQ2. How is the overall performance of the MEGP using the best prediction approach compared with the standard GP for mill throughput prediction?

To achieve the aim of this study, a data set comprised of 20,161 industrial records from a gold mining complex in Western Australia is used. Based on expert knowledge, the influential inputs from the available data set are selected as mill power consumption, mill turning speed, inlet water, and input particle size.

The remainder of the paper is organised as follows. The subsequent section is dedicated to data set and presenting essential statistics of variables along with performing some data pre-possessing steps to make data ready to be used by prediction methods. GP algorithm and newly proposed GP method (MEGP) will be discussed in section III. Results and discussions are presented in section IV, and finally, section V summarises findings of this research.

## II. DATA SET AND PRE-PROCESSING

In the original data set, certain erroneous measurements were identified, such as negative values for mill turning speed and percentage data exceeding 100%. These erroneous data were eliminated. For removing outliers, we relied on expert's knowledge, as sometimes outliers indicate regular operation of the system that may recur in the future and should be considered in model training. According to experts' recommendation, only power outliers were eliminated as indicating non-operational periods of the mill. After removing erroneous data and outliers, the number of data points reduced to 19,724. TABLE I displays the descriptive statistics of the cleaned data. In this table, %PL variables indicates the percentage of particles with a special size, including less than 19 mm, 19-53 mm, and 53-300 mm. To gain a deeper understanding of the data, we utilize the three-dimensional projection using t-SNE methodology [30] and the correlation heatmap, as depicted in Fig. 1.

As can be seen, in Fig1(a), no outlier is observed after data cleaning. The correlation heatmap reveals that mill throughput is predominantly influenced by mill weight and inlet water factors.

## III. METHODS

### A. Genetic programming

Genetic programming (GP) developed by Koza [12] is a highly efficient approach inspired by biological evolution and natural selection to derive computer programs capable of solving a remarkable range of diverse problems. GP can be utilised for regression purpose, through exploring accurate mathematical equations between input and output variables. To initiate the process, a population of random equations in the form of expression trees (ET) is generated.

Each ET consists of functions as branches and a terminal set comprising input variables and constants. The function set can include basic mathematical operators (i.e., +, -, ×, /), Boolean logic functions (i.e., AND, OR, NOT), trigonometric functions (i.e., sin, cos), as well as user-defined functions. In the next step, fitness value for each created equation is calculated based on the defined measure and similar to the principles of Darwinian evolution, ETs with the highest performance are selected to create the next generation of ETs through applying crossover and mutation operators. During crossover, two well-performing ETs are selected as parents and mated to create a new offspring. Mutation is applied to prevent getting stuck in local optima by introducing variations in ETs and exploring new areas of the search space.

TABLE I. BASIC DESCRIPTIVE STATISTICS OF THE DATA

| Variable | Unit | Min | Max | Mean | SD |
|---|---|---|---|---|---|
| Mill throughput | t/h | 762.24 | 1616.98 | 1255.83 | 89.42 |
| Mill power draw | kW | 12000.24 | 14359.50 | 13172.68 | 431.08 |
| Mill turning speed | rpm | 8.34 | 10.31 | 9.99 | 0.36 |
| Inlet water | m3/h | 200.35 | 482.85 | 402.79 | 39.42 |
| %PL 0-19 | % | 24 | 47 | 33.98 | 2.93 |
| %PL 19-53 | % | 20 | 36 | 28.74 | 1.88 |
| % PL 53-300 | % | 22 | 56 | 37.26 | 4.47 |

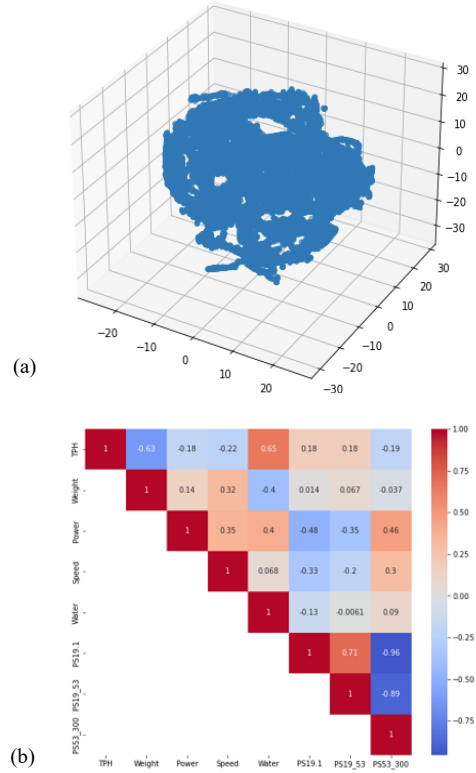

Fig. 1. t-SNE Projection (a) and (b) correlation heatmap

The process of creating new generation continues until reaching the termination criterion, which can be defined as the maximum number of generations, a predetermined execution time or a quality threshold.

### B. New proposed GP method (MEGP)

This section provides details for new GP method developed in this study. The fundamental concept for the new method (MEGP) as outlined in Algorithm 1 is the creation of several equations through the application of GP to the training data set, and then applying different approaches to utilise these equations for making predictions on the test data. These steps can be categorised into two main categories of data clustering and prediction.

#### 1) Data Clustering

The aim of the clustering step is exploring subgroups of train data that can be accurately modelled by equations derived using GP. To achieve this, GP is executed separately 30 times. The most accurate equation out of these runs is identified and used for separating the first cluster of input data. For this purpose, prediction error for each train data point using the best-found equation is obtained and the train data points with prediction errors lower than a predetermined error threshold ($\varepsilon$), will be considered as cluster 1. Moreover, the most accurate equation is retained for the prediction step. In this study, the median of absolute prediction errors for train data is determined as the error threshold. The median is preferred for its robustness to

**Algorithm 1** Proposed GP Algorithm (MEGP)

**0.1. Data Clustering**

**Input:** Training dataset D, Predetermined limit for remaining training data N, Number of runs for GP R
**Output:** List of clusters for categorized data C, Best models for data clusters $M_{best}$

1: **Initialize Variables:**
2: $C \leftarrow$ Empty list.
3: $M_{best} \leftarrow$ Null.
4: $RMSE_{best} \leftarrow \infty$.
5: **for** $r$ in 1 to $R$:
6:     Run GP on D to create a model: $M_r$.
7:     Calculate RMSE for the current model: $RMSE_r$.
8:     **if** $RMSE_r < RMSE_{best}$:
9:         $M_{best} \leftarrow M_r$     ▷ Update best model.
10:        $RMSE_{best} \leftarrow RMSE_r$     ▷ Update minimum RMSE.
11:    **end if**
12: **end for**
13: **for** $i$ in 1 to $N_D$:     ▷ For all training data points.
14:     Calculate prediction error using $M_{best}$: $E_i$.
15: **end for**
16: Calculate median of absolute prediction errors (MAPE) as $\epsilon$
17: **for** $i$ in 1 to $N_D$:
18:     **if** $|E_i| < \epsilon$:
19:         Assign data points to a new cluster and remove from D
20:     **end if**
21: **end for**
22: Add the new cluster to C.
23: **if** remaining data points in D > N:
24:     Go back to step 5 with the remaining D.
25: **end if**
26: **Finish Clusteirng:**
27: **Return** C and $M_{best}$ for the prediction step

**0.2. Prediction**

**Input:** Test data T, List of clusters C, Best models for data clusters $M_{best}$,
**Output:** Predictions for test data using different approaches

28: **for** each test data point i in T:
29:     **for** each data cluster j in C:
30:         m ← Number of data clusters
31:         $n_j$ ← Number of data points in each data cluster j
32:         $p_{ij}$ ← Predicted value of data point i using the best model for data cluster j ($M_{best}$)
33:         $d_{ij}$ ← normalized minimum distance data point i from each data cluster j using four distance measures
34:     **end for**
35:     **if** prediction Method is GP-best-cl:
36:         best cluster ← Find the cluster with the minimum distance
37:         $p_i$ ← Evaluate Equation ($M_{best}$, [best cluster], i)
38:     **else if** prediction Method is GP-sim-avg:
39:         $p_i$ ← Evaluate Equation 1 ($p_{ij}$,m)
40:     **else if** prediction Method is GP-Weighted average n:
41:         $p_i$ ← Evaluate Equation 2 ($p_{ij}$, $n_j$, m)
42:     **else if** prediction Method is GP-Weighted average d:
43:         $p_i$ ← Evaluate Equation 3 ($p_{ij}$, $d_j$, m)
44:     **else if** prediction Method is GP-Weighted average nd:
45:         $p_i$ ← Evaluate Equation 4 ($p_{ij}$, $d_{ij}$, $n_j$, m)
46:     **end if**
47: **end for**
48: **Return predicted values for test data points ($p_i$)**

outliers, ensuring that extreme values in the prediction errors do not influence the threshold determination. This characteristic aligns well with scenarios where the dataset may contain instances with unusually high or low errors. Moreover, the median's resilience to skewed distributions makes it a reliable measure of central tendency in cases where other statistics, such as the mean, might be sensitive to such deviations.

In the next step, data points allocated to cluster 1 are removed from train data and GP is executed again 30 time using all remained trin data. The second cluster is created in the similar way and this iterative process continues until the number of remained train data is less than the predetermined limit (N) which is selected as 1% of train data. At the end of the data clustering step, multiple categories of training data, each associated with a mathematical equation capable of accurately modelling the respective cluster, are obtained.

*2) Prediction*

In order to predict mill throughput as the output variable for the test data, we employ five different approaches, outlined as follows:

I) Best cluster (GP-best-cl): Calculating the minimum distance between each test data point and all data points within each cluster of the train data and selecting the cluster with the shortest distance as the best cluster. Utilising the equation associated with the best cluster to calculate prediction for the test data.

II) Simple average (GP-sim-avg): Simple average of all equations associated with all data clusters.

III) Weighted average n (GP-w-avg(n)): Weighted average of all equations based on the number of data points in each data cluster.

IV) Weighted average d (GP-w-avg(d)): Weighted average of all equations based on the minimum distance from each data cluster.

V) Weighted average n and d (GP-w-avg(nd)): Weighted average of all equations based on the number of data points in each data cluster and the minimum distance from them.

The prediction value for the i[th] test data point, $p_i$, using the aforementioned methods can be calculated using the following equations.

$$p_{i-(GP-Sim-avg)} = \frac{\sum_{j=1}^{m} p_{ij}}{m} \quad (1)$$

$$p_{i-(GP-w-avg(n))} = \frac{\sum_{j=1}^{m} n_j p_{ij}}{\sum_{j=1}^{m} n_j} \quad (2)$$

$$p_{i-(GP-w-avg(d))} = \frac{\sum_{j=1}^{m} (1-d_j) p_{ij}}{\sum_{j=1}^{m} (1-d_j)} \quad (3)$$

$$p_{i-(GP-w-avg(nd))} = \frac{\sum_{j=1}^{m} n_j (1-d_j) p_{ij}}{\sum_{j=1}^{m} n_j (1-d_j)} \quad (4)$$

where $p_{ij}$ represents predicted values of the i$^{th}$ test data point using the j$^{th}$ equation. m is the total number of clusters. $n_j$ and $d_j$ indicates the number of data points in each cluster, and the normalized minimum distance from each cluster, respectively. The distance between data points is computed using four different measures including Euclidean, Manhattan, Chebyshev, and Cosine distances [31]. Each measure employs a specific equation to quantify the distance between two points as follows:

$$\text{Euclidean distance } (p, q) = \sqrt{\sum_{i=1}^{n}(p_i - q_i)^2} \quad (5)$$

$$\text{Manhattan distance } (p, q) = \sum_{i=1}^{n}|p_i - q_i| \quad (6)$$

$$\text{Chebyshev distance } (p, q) = \max_i |p_i - q_i| \quad (7)$$

$$\text{cosine distance } (p, q) = 1 - \frac{p \cdot q}{\|p\| \|q\|} \quad (8)$$

In these equations, p and q are two data points in n-dimensional space with Cartesian coordinates (p$_1$, p$_2$, …, p$_n$) and (q$_1$, q$_2$, …, q$_n$). Among these distance measures, the Euclidean distance is a widely used metric, calculates the straight-line distance between two points and is sensitive to differences in all dimensions. Manhattan distance, also known as City Block distance, measures the distance by moving along the axes and summing the absolute differences along each dimension. Chebyshev distance, also called the Maximum Value distance, considers only the largest difference in any dimension. Cosine distance is a measure of similarity between two vectors. It calculates the cosine of the angle between the two vectors in a multi-dimensional space.

Using different distance measures enables a thorough evaluation of the proposed methods. It helps us to understand how each measure influences accuracy and model suitability for our dataset. This approach ensures a robust assessment and helps us to choose the most suitable distance measure for each new GP variant.

### C. Model setting and validation

*1) Model parameters*

Before implementing GP, some parameters need to be set. Parameters considered in this study are provided in Table II. The population size and maximum number of generations are determined based on recommended values in [32]. For function set, we have extended the default operator set by incorporating square root (sqrt) and logarithmic (log) functions. For terminal set, in addition to input variables, we considered intercepts from range (-1000, 1000) to provide an appropriate range of intercepts for equations.

The remaining inputs are based on predefined values suggested in the original code. The Python package "gplearn" is utilized in this research as it is an efficient GP implementation that can provide the required customization options [33].

TABLE II. GP PARAMETERS

| Parameters | Values |
|---|---|
| Function set | {+, -, *, / , sqrt, log} |
| Terminal set | Input variables and constants as random numbers from range (-1000, 1000) |
| Initialization method | Half and half |
| Maximum tree depth | Range (2,6) |
| Population size | 200 |
| Maximum number of generations | 500 |
| Fitness function | Mean absolute error (MAE) |
| Selection method | Tournament |
| Crossover probability | 0.9 |
| Mutation probability | 0.01 |
| Stopping criterion | Maximum generation |

*1) Model validation*

In order to evaluate the predictive performance of prediction models, a commonly employed technique is k-fold cross-validation [34]. However, when dealing with time series data, the conventional k-fold cross-validation approach is unsuitable as it disregards the time sequence. In other words, in time series, future data cannot be used to predict past data. To address this challenge, we utilize an adapted version of k-fold cross-validation for time series known as rolling-origin-recalibration evaluation [35], referred as rolling-based cross validation in this study, as depicted in Fig. 2.

In this approach data is divided into k folds. The model is trained using the first fold and tested on the subsequent fold. After each evaluation, the window slides forward, creating a new training set including both pervious train and test folds and using the subsequent fold for validation.

The model is retrained on the updated training set and evaluated again. This process continues until the entire time series has been utilised for training and testing. In this research, rolling-based cross validation method with k=10 folds is performed.

### IV. RESULTS AND DISCUSSIONS

To compare the performance of different methods, we utilise the mean absolute error (MAE) index, computed as follows:

$$MAE_{s_i} = \frac{1}{N_{s_i}} \sum_{i=1}^{N_{s_i}} |y_i - y'_i| \quad (9)$$

Where $y_i$ and $y'_i$ represents the actual and predicted mill throughput values for the test data points in the split $s_i$ and $N_{s_i}$ denotes the number of test data points in this split.

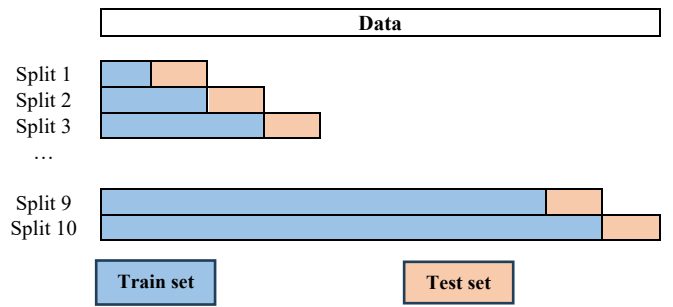

Fig. 2. rolling-based cross-validation for time series

TABLE III presents the mean absolute error (MAE) results obtained by applying the standard GP (GP-std) and new proposed GP (MEGP) using different prediction approaches to various data splits employing different distance measures. The best prediction results for each data split are highlighted.

The first research question aims at identifying the best prediction approach for MEGP (RQ1). To achieve this objective, first the best distance measure for distance-dependent methods is identified and then each prediction approach using the best distance measure is compared. considering results in TABLE III, we identified Manhattan and Euclidean distance as the optimal measures for the MEGP (GP-w-avg(d)) and MEGP (GP-w-avg(nd)) methods, respectively. These distance measures resulted in the least average error for all data splits for their respective methods. Their results either ranked as the best or closely matched the best results for data splits individually. Regarding the GP-best-cl method, Euclidean distance exhibited the lowest average error. However, variations were observed when considering prediction errors across different data splits. Hence, for the final comparison, we will include all different distance measures for this method. The MAE prediction results for GP-std, and MEGP with different prediction approaches as GP-sim-avg, GP-best-cl using Euclidean, Manhattan, and Cosine distance measures, GP-w-avg(d) using Manhattan distance, GP-w-avg-n, and GP-w-avg(nd) using Euclidean distance for different data splits is represented in Fig. 3.

data splits 4, 6 and 8. Its average MAE is 56.54 which is higher than the initial GP's MAE of 51.58. Overall, its performance is not better than standard GP. This method does not utilise any properties of data clusters for making final predictions and could not outperform the standard GP.

MEGP (GP-best-cl) solely relies on the equation associated with the cluster displaying the shortest distance to each test data point, without considering other equations or the number of data points in each cluster. As a result, this approach yields diverse outcomes. For instance, in data split 9, MEGP (GP-best-cl), utilising all three distance measures, achieves notably superior results, surpassing the initial GP's performance by a considerable margin. Even in data split 7, this method, when employing Euclidean distance, achieves the most accurate prediction results among all eight methods. However, in contrast, for data splits 1 and 6, its prediction errors using all three distance measures are notably high.

This result highlights that relying solely on distance and utilising just one equation for prediction can introduce significant prediction variability, leading to both highly favorable and unfavorable outcomes, thereby reducing stability in the results.

Among the three remained weighted approaches including MEGP(GP-w-avg(n)), MEGP(GP-w-avg(d)), and MEGP (GP-w-avg(nd)), method MEGP (GP-w-avg(d)), showed slightly superior results only for data split 4.

Comparing the performance of MEGP (GP-w-avg(n)) and MEGP (GP-w-avg(nd)) reveals that MEGP (GP-w-avg(nd)) outperforms MEGP (GP-w-avg(n)), with an average MAE of 46.04 compared to 46.62 for MEGP(GP-w-avg(n)). These findings provide compelling evidence that incorporating more information about data clusters leads to the most accurate predictions.

Notably, MEGP (GP-w-avg(nd)) yields the least prediction error for 6 out of 10 data splits, and for the 2 other data splits (splits 2 and 5), its results are in close proximity to the best outcomes. Consequently, this method achieves the best or nearly best results for 8 out of 10 data splits, highlighting its exceptional performance. Significant improvements were observed, particularly for data splits 4, 8, and 9.

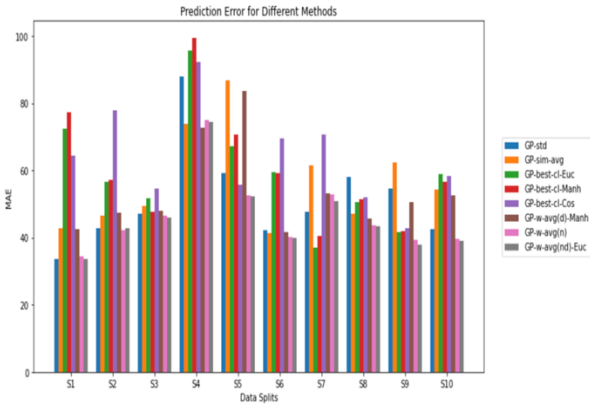

Fig. 3. MAE results per data splits

TABLE III. MAE VALUES PER DATA SPLITS USING STANDARD AND NEW GP VARIANTS WITH DIFFERENT DISTANCE MEASURES

| Methods | Distance measure | Data Splits | | | | | | | | | | Mean | SD |
|---|---|---|---|---|---|---|---|---|---|---|---|---|---|
| | | S1 | S2 | S3 | S4 | S5 | S6 | S7 | S8 | S9 | S10 | | |
| GP-std | - | 33.65 | 42.88 | 47.17 | 87.84 | 59.14 | 42.26 | 47.63 | 58.13 | 54.65 | 42.42 | 51.58 | 15.01 |
| MEGP(GP-sim-avg) | - | 42.71 | 46.37 | 49.27 | 73.86 | 86.77 | 41.33 | 61.45 | 47.02 | 62.2 | 54.39 | 56.54 | 14.70 |
| MEGP (GP-w-avg(n)) | - | 34.32 | 42.21 | 46.38 | 74.85 | 52.61 | 40.27 | 52.73 | 43.68 | 39.36 | 39.74 | 46.62 | 11.50 |
| MEGP (GP-best-cl) | Euc | 72.46 | 56.59 | 51.65 | 95.80 | 67.20 | 59.32 | **37.14** | 50.50 | 41.65 | 58.78 | 59.11 | 16.71 |
| | Manh | 77.26 | 57.11 | 47.7 | 99.49 | 70.62 | 59.22 | 40.51 | 51.45 | 41.9 | 56.68 | 60.19 | 18.02 |
| | Cheb | 72.46 | 56.59 | 51.65 | 95.80 | 67.20 | 59.32 | **37.14** | 50.50 | 41.65 | 58.78 | 59.11 | 16.71 |
| | Cos | 64.41 | 77.87 | 54.49 | 92.35 | 55.57 | 69.43 | 70.56 | 52.11 | 42.8 | 58.17 | 63.78 | 14.35 |
| MEGP (GP-w-avg(d)) | Euc | 41.84 | 47.51 | 48.04 | **72.36** | 84.86 | 41.56 | 53.30 | 45.61 | 50.48 | 53.29 | 53.89 | 13.96 |
| | Manh | 42.42 | 47.32 | 47.86 | 72.56 | 83.7 | 41.54 | 53.04 | 45.77 | 50.42 | 52.67 | 53.73 | 13.67 |
| | Cheb | 41.84 | 47.51 | 48.04 | **72.36** | 84.86 | 41.56 | 53.30 | 45.61 | 50.48 | 53.29 | 53.89 | 13.96 |
| | Cos | 42.44 | 46.89 | 48.86 | 72.82 | 84.46 | 41.48 | 58.8 | 44.9 | 59.55 | 53.82 | 55.40 | 14.01 |
| MEGP (GP-w-avg(nd)) | Euc | **33.62** | 42.73 | **46.06** | 74.51 | 52.28 | **39.94** | 50.8 | **43.41** | **38.02** | **39.05** | **46.04** | 11.52 |
| | Manh | 33.87 | 42.37 | 46.15 | 73.9 | **52.12** | 40.01 | 51.18 | 43.72 | 38.41 | 39.16 | 46.09 | 11.30 |
| | Cheb | **33.62** | 42.73 | **46.06** | 74.51 | 52.28 | **39.94** | 50.80 | **43.41** | **38.02** | **39.05** | **46.04** | 11.52 |
| | Cos | 33.93 | **42.16** | 46.31 | 74.42 | 52.24 | 40.23 | 52.73 | 43.95 | 38.4 | 39.58 | 46.40 | 11.49 |

Considering all discussions, we can conclude that MEGP (GP-w-avg(nd)) using Euclidean distance which utilises all obtained equations and considers both distances to data splits and the number of data points in each split is the best performing method among all new GP variants.

Having discussed the best-performing approach for MEGP (RQ1), our focus is now to compare the predictive performance of the MEGP with the best prediction approach with the standard GP (RQ2).

We calculate the improvement percentages for each data split using the following formula:

$$\% \ improvement = \left( \frac{MAE_{GP-std} - MAE_{best}}{MAE_{GP-std}} \right) * 100 \quad (10)$$

where $MAE_{GP-std}$ and $MAE_{best}$ represent prediction errors of the standard GP and the best performing model, respectively.

TABLE IV shows the MAE results of the standard GP and the best-performing MEGP variant using Euclidean distance along with the Wilcoxon rank-sum test statistical results and improvement percentages. These results are also depicted in Fig. 4.

As it is evident, for all data splits, except for split 7, this new GP variant could outperform the standard GP with statistically significant improvements observed in S3, S4, S5, S6, S8, S9, and S10, as indicated by p-values below the selected significance level of 0.05. Specifically, this new GP variants exhibited superior performance for data splits 4, 8, and 9, showing improvements of 15.18%, 25.32%, and 30.43% respectively compared to the standard GP. The average MAE for the standard GP was 51.58, while the MEGP (GP-w-avg(nd)) achieved a significantly lower average MAE of 46.04, representing a remarkable 10.74% improvement. Furthermore, the statistical significance of this superiority becomes evident across most large data splits, progressing from split 1 to split 10. These findings underscore the suitability of adopting the new GP method, particularly in the context of large datasets. Overall, these findings confirm the superiority of the the MEGP (GP-w-avg(nd)) using Euclidean distance measures compared with the standard GP approach (addressing RQ2). New GP method developed in this study offers enhanced accuracy for mill throughput prediction.

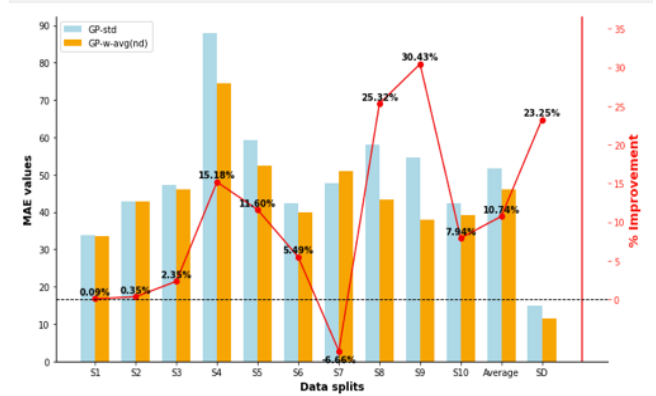
Fig. 4. MAE results of the standard GP and the GP-w-avg(nd) with the improvement percentages

Further analysis of the results depicted in Fig. 3 reveals that certain data splits, such as splits 4 and 5, present greater challenges for most methods, resulting in higher prediction errors. Furthermore, it can be generally observed that as we progress from split 1 to split 10 and increase the size of the training data, MEGP, particularly MEGP (GP-w-avg(nd)) using Euclidean distance, demonstrate more significant improvements, which is reasonable. As the training data volume grows, the practice of partitioning them into distinct clusters, which can be more precisely modelled with separate equations, becomes more meaningful. This approach is likely to result in improved predictions for more data points.

## V. CONCLUSION

This study focuses on GP enhancement for more accurate prediction of SAG mill throughput. The newly developed GP method (MEGP) utilise a clustering technique to partition the training data into distinct groups, in such a way that each cluster can be precisely modelled by an individual equation.

Subsequently, five distinct approaches are introduced to leverage these identified clusters and equations for predicting the test data. This study utilises four distance measures to assess their effects on the model's performance. In order to evaluate the effectiveness of MEGP, a data set comprising 20,161 industrial records is utilised.

TABLE IV. MAE VALUES OF DATA SPLITS USING GP_STD AND THE BEST NEW GP VARIANT AND WILCOXON RANK-SUM TEST STATISTICAL RESULTS TO INDICATE THE SIGNIFICANT DIFFERENCE OF THESE TWO GP VARIANTS

| Methods | Data Splits | | | | | | | | | | Mean | SD |
| --- | --- | --- | --- | --- | --- | --- | --- | --- | --- | --- | --- | --- |
| | *S1* | *S2* | *S3* | *S4* | *S5* | *S6* | *S7* | *S8* | *S9* | *S10* | | |
| GP-std | 33.65 | 42.88 | 47.17 | 87.84 | 59.14 | 42.26 | 47.63 | 58.13 | 54.65 | 42.42 | 51.58 | 15.01 |
| MEGP (GP-w-avg(nd))_Euc | 33.62 | 42.73 | 46.06 | 74.51 | 52.28 | 39.94 | 50.8 | 43.41 | 38.02 | 39.05 | 46.04 | 11.52 |
| % improvement | 0.09% | 0.35% | 2.35% | 15.18% | 11.60% | 5.49% | -6.66% | 25.32% | 30.43% | 7.94% | 10.74% | 23.25% |
| P-value | 8.35E-01 | 2.01E-01 | 4.37E-02 | 1.34E-03 | 5.01E-03 | 3.47E-02 | 1.04E-02 | 9.08E-04 | 1.78E-05 | 1.82E-03 | - | - |

Given the time series nature of the data set, a rolling-based cross validation method, which is a special type of common k-fold cross validation for time series data, is employed.

Among different prediction approaches of MEGP, the comparison results highlight the superiority of the MEGP(GP-w-avg(nd)) utilising the Euclidean distance. This variant incorporates all explored equations for all data clusters in its prediction process and use both the distance to clusters and the number of data points in each cluster as weighting factors for prediction results. Consequently, this approach could outperform standard GP with a considerable enhancement of 10.74% in prediction accuracy and 23.25% decrease in the standard deviation of the results. This method could attain the better results for 9 data splits with statistically significant improvements observed in 7 data splits.

In conclusion, the results of this study demonstrate the reliability and efficacy of MEGP for predicting SAG mill throughput. Moreover, this method can be utilised in other modelling problems. Exploring the impact of error threshold (epsilon) and minimum number of data points for creating clusters (N) on the performance of these variants, can be considered as a direction for future research. Furthermore, another direction for future research involves integrating these approaches with other machine learning methods to enhance their prediction accuracy.

ACKNOWLEDGEMENTS

This research was supported by the Australian Research Council Integrated Operations for Complex Resources Industrial Transformation Training Centre (project number IC190100017) and funded by universities, industry and the Australian Government.